\documentclass{article}

\usepackage[preprint]{neurips_2025}

\usepackage[utf8]{inputenc} 
\usepackage[T1]{fontenc}    
\usepackage{hyperref}       
\usepackage{url}            
\usepackage{booktabs}       
\usepackage{amsfonts}      
\usepackage{nicefrac}       
\usepackage{microtype}      
\usepackage{xcolor}         
\usepackage{natbib}
\usepackage{enumitem}
\usepackage{amssymb}
\usepackage{amsmath}
\usepackage{graphicx}
\usepackage{algorithm}
\usepackage{algorithmic}
\usepackage{adjustbox}
\usepackage{graphicx}
\usepackage{subcaption}
\usepackage{hyperref}

\usepackage{inconsolata}

\usepackage[most]{tcolorbox}  
\tcbset{
  mypromptbox/.style={
    colback=gray!10,    
    colframe=gray!60,    
    boxrule=0.5pt,
    arc=2pt,
    left=3pt,
    right=3pt,
    top=4pt,
    bottom=4pt,
    enhanced,
    sharp corners,
    fonttitle=\bfseries,
    breakable,
  }
}

\title{LLM-Guided Reinforcement Learning: Addressing Training Bottlenecks through Policy Modulation}

\author{
  Heng Tan \quad Hua Yan \quad Yu Yang \\
  Lehigh University \\
  \texttt{\{het221, huy222, yuyang\}@lehigh.edu}
}

\begin{document}

\maketitle

\begin{abstract}
\par While reinforcement learning (RL) has achieved notable success in various domains, training effective policies for complex tasks remains challenging. Agents often converge to local optima and fail to maximize long-term rewards. Existing approaches to mitigate training bottlenecks typically fall into two categories: (i) Automated policy refinement, which identifies critical states from past trajectories to guide policy updates, but suffers from costly and uncertain model training; and (ii) Human-in-the-loop refinement, where human feedback is used to correct agent behavior, but this does not scale well to environments with large or continuous action spaces.
In this work, we design a large language model-guided policy modulation framework that leverages LLMs to improve RL training without additional model training or human intervention. We first prompt an LLM to identify critical states from a sub-optimal agent’s trajectories. Based on these states, the LLM then provides action suggestions and assigns implicit rewards to guide policy refinement. Experiments across standard RL benchmarks demonstrate that our method outperforms state-of-the-art baselines, highlighting the effectiveness of LLM-based explanations in addressing RL training bottlenecks.

\end{abstract}

\section{Introduction}

\par While Reinforcement Learning (RL) has shown strong potential in domains such as simulation games~\cite{cai2023imitation,liuutility}, robotics~\cite{yin2024offline, shridhar2023perceiver}, and traffic control systems~\cite{tan2023joint, huang2024uncertainty}, training an optimal policy for complex tasks still remains challenging. RL training often encounters bottlenecks, such as getting stuck in local optima, which prevents the agent from achieving the maximum cumulative reward. Re-training policies from scratch to overcome these issues is often computationally expensive and time-consuming~\cite{agarwal2022reincarnating, vinyals2019grandmaster}, which highlights the need for more efficient approaches to address training bottlenecks in RL.

\par Existing approaches for addressing reinforcement learning (RL) training bottlenecks can be broadly divided into two categories.
(i) Automated policy refinement with explanation:
These methods begin with a non-optimal RL policy and train a separate network to identify critical states—states that strongly influence long-term rewards—as a form of explanation. Based on these identified states, different strategies are applied: recommending actions~\cite{guo2021edge}, preventing the agent from entering misleading state-action pairs~\cite{liuutility}, or encouraging exploration at those states~\cite{cheng2023statemask, cheng2024rice, yu2023airs}. However, these approaches face two key challenges: (1) definitions of critical states vary significantly across methods, and (2) training a network to reliably identify them introduces uncertainty and can be computationally expensive for complex tasks.
(ii) Human-in-the-loop RL refinement:
These methods rely on human experts to identify important time steps and either provide corrected actions or restrict certain actions at those points~\cite{van2022correct, guan2021widening}. While effective in smaller settings, such approaches become less practical as the action space grows, due to increased cognitive load and scalability issues.

\par Recent advances in reinforcement learning (RL) augmented by large language models (LLMs) have demonstrated the effectiveness of LLMs in natural language understanding, reasoning, and task planning, positioning them as a promising alternative to humans for guiding policy learning in complex tasks. For example, LLM can serve as reward designers, providing implicit or explicit rewards~\cite{kwon2023reward, wu2023read, yu2023language}, or as decision makers, offering suggested action candidates or prior policies~\cite{li2022pre, ahn2022can, yao2020keep} to guide RL policy learning. However, directly involving LLMs in the policy retraining process for complex tasks is prohibitively expensive in terms of both computational time and financial cost; for instance, in environments such as MuJoCo, the training process often requires over one million episodes on average~\cite{cheng2023statemask}. To address this challenge, our work explores how LLM-generated explanations can be used to mitigate RL training bottlenecks and enhance policy performance, without requiring full retraining.

\par We introduce ULTRA, a framework of \underline{u}tilizing large \underline{l}anguage model-based explana\underline{t}ion to address the training bottlenecks for \underline{r}einforcement le\underline{a}rning. Our method consists of two main components: (i) \textbf{Identification: }We first leverage an LLM to identify the critical states that are most influential to future rewards. Specifically, we collect historical trajectories from a non-optimal RL policy and provide them to the LLM, along with environment information and natural language-based identification instructions.
The LLM processes this input to pinpoint the critical states within the trajectories. (ii) \textbf{Refinement: }Based on the identified critical states, we integrate the LLM into the agent’s Markov Decision Process (MDP) to refine its policy. We explore three strategies:
(a) Given the identified critical states, we prompt the LLM to suggest optimal actions. These actions are stored in a lookup table, which the RL agent consults during policy fine-tuning.
(b) We prompt the LLM to analyze the agent’s effective and sub-optimal actions at the critical states, producing explanations that guide reward shaping.
(c) We combine both (a) and (b) to jointly leverage LLM-suggested actions and explanation-driven reward signals during the fine-tuning process.

\par In summary, this paper makes the following contributions.
\begin{itemize}[leftmargin=*]
\item We introduce a framework that leverages LLM-based explanations to address training bottlenecks for RL. Specifically, we use an LLM to identify critical states from the historical trajectories of a non-optimal agent and explore three strategies for improving policy performance.
\item We evaluate our method across five environments, ranging from a simple sparse-reward game (e.g., Pong) to complex dense-reward tasks (e.g., MuJoCo [Ant]). Experimental results demonstrate that our approach outperforms state-of-the-art baselines.
\item We assess the explanation fidelity of different models and analyze how varying explanation affect the RL policy refinements. The experimental results show that our method achieves superior performance compared to existing baselines.
\end{itemize}

\section{Related Work}
\subsection{Policy Refinement}

Reinforcement learning (RL) often suffers from inefficient training due to challenges such as sparse rewards, suboptimal exploration, and unstable policy updates. To address these bottlenecks, many existing methods focus on policy refinement—the process of improving a non-optimal policy by leveraging external signals or structural insights, which can be broadly classified into two categories: (i) automated policy refinement with explanation, and (ii) human-in-the-loop RL refinement.

\par(i) Automated Policy Refinement with Explanation.
Given a non-optimal RL policy, some approaches first discuss how the critical states are defined, and then train an auxiliary network to identify critical states as a form of explanation for the further policy refinement: perturbing the agent's actions~\cite{guo2021edge,liuutility}, or encouraging targeted exploration~\cite{cheng2023statemask, cheng2024rice, yu2023airs}. For example, StateMask~\cite{cheng2023statemask} learns to mask non-critical states while preserving the agent’s performance, revealing key decision points that impact the final reward. This insight supports downstream applications such as adversarial testing and error correction.
UTILITY~\cite{yu2023airs} adopts a two-level explainability framework to detect agent mistakes and reformulate policy refinement as a bi-level optimization problem, incorporating reward shaping and constrained learning.
However, these methods are sensitive to how critical states are defined, and the accuracy of the trained network significantly affects downstream performance. Moreover, training such networks for complex tasks is often computationally expensive and time-consuming.

\par(ii) Human-in-the-Loop RL Refinement.
Other methods incorporate expert feedback, where humans identify key time steps and either suggest improved actions or restrict undesirable ones~\cite{van2022correct, guan2021widening, laleh2024survey, metz2024mapping}.
For example, Van der Pol et al.~\cite{van2022correct} use human feedback after agent failures to create action-level shields that constrain high-level decisions, improving learning efficiency and policy robustness.
Zhang et al.~\cite{guan2021widening} propose a framework that combines binary human feedback with visual explanations, enabling context-aware data augmentation that preserves human-salient features.
Despite their effectiveness, these approaches face scalability issues in environments with large or continuous action spaces, as human input becomes increasingly difficult to collect and apply efficiently.

\subsection{LLM-enhanced  Reinforcement Learning}
Recent advances in reinforcement learning augmented with large language models have demonstrated the potential of LLMs in natural language understanding, reasoning, and task planning. These capabilities position LLMs as a promising alternative to humans for guiding policy learning in complex tasks. 
They can serve as reward designers and decision makers, depending on how their outputs are used to guide the agent’s learning process.
(i) LLMs as reward designers. LLMs can generate implicit or explicit reward signals conditioned on natural language task descriptions, agent behaviors, or environmental context~\cite{kwon2023reward, wu2023read, yu2023language}. For example, Yu et al.~\cite{yu2023language} propose a language-guided reward shaping framework, where LLMs translate high-level language instructions or corrections into parameterized reward functions. These rewards are optimized in real time using a model-predictive controller, enabling interactive and flexible robot behavior generation across diverse tasks. 
(ii) LLMs as decision makers. LLMs can propose candidate actions, predict the next best step in a task, or offering prior policy structures based on language instructions or few-shot examples~\cite{li2022pre, ahn2022can, yao2020keep, huang2022inner}. For example, Ahn et al.~\cite{ahn2022can} introduce Say-Can, a framework that combines LLM-generated high-level action suggestions with pretrained low-level skills, enabling robots to follow abstract natural language instructions while ensuring real-world feasibility through value-based grounding.
However, directly involving LLMs in the policy retraining process for complex tasks is prohibitively expensive in terms of both computational time and financial cost.

\section{Methodology}
\label{sec:method}
 \begin{figure}[h]

    \centering
    \vspace{-10pt}
    \includegraphics[width=0.8\linewidth, keepaspectratio=true]{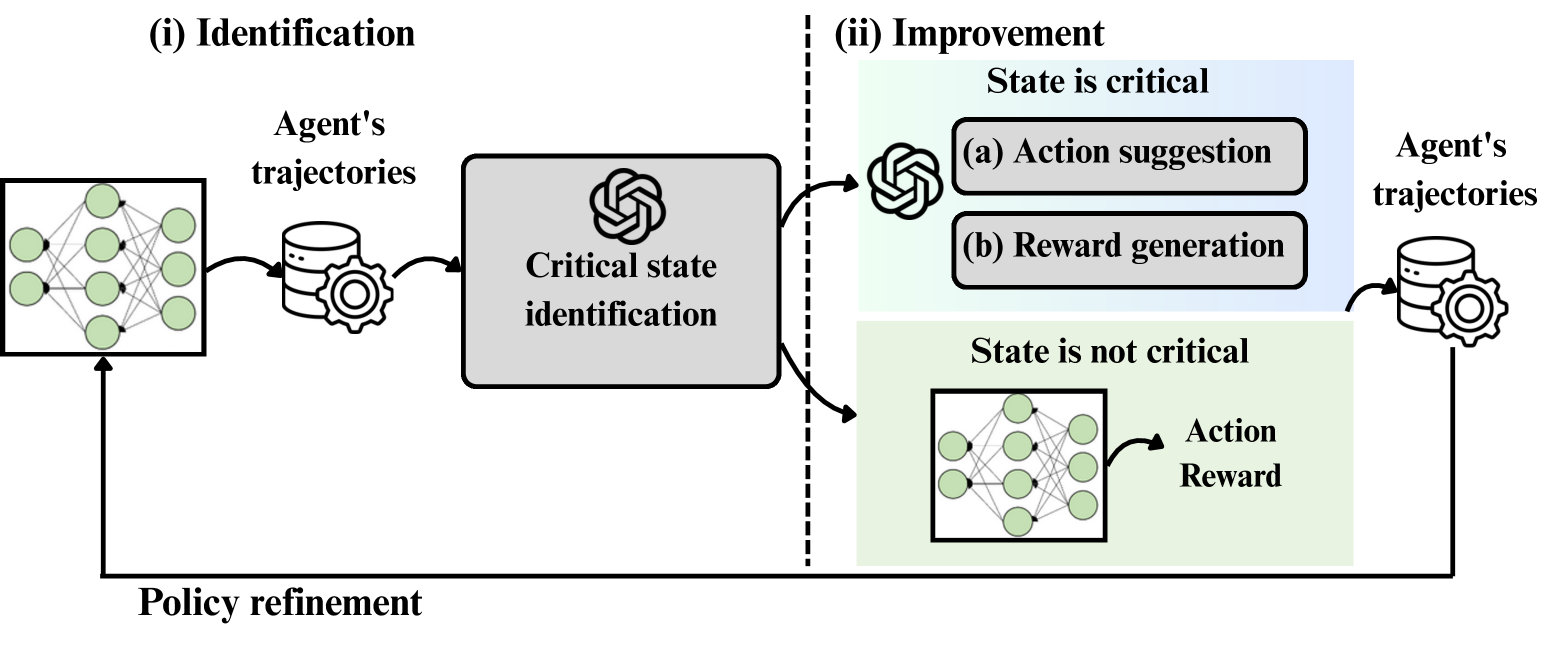}
    \vspace{-10pt}
    \caption{An overview of our framework. (i) Identification: we collect trajectories from a suboptimal RL policy, convert them into natural language with environment context, and prompt an LLM to identify critical states in each episode. (ii) Improvement: after critical states are identified, the agent follows its original policy at non-critical states, while at critical states, it adopts the actions suggested by the LLM and receives the corresponding LLM-generated rewards. The trajectories generated after (i) and (ii) serve as training data for further policy updates. 
}

    \label{fig:model}

\end{figure}

\subsection{Problem Setup}
\par The RL agent's decision making is based on the Markov Decision Process (MDP), which is defined as a tuple $G = \{\mathcal{S}, \mathcal{A}, \mathcal{R}, \mathcal{P}, \gamma\}$. $\mathcal{S}$ represents the agent's state set. $\mathcal{A}$ denotes the agent's action set. 
$\mathcal{R}$ is the reward function. 
$\mathcal{P}$ denotes the transition probability function. $\gamma$ is the discount factor.
Given the RL policy $\pi$, its goal is to maximize the cumulative reward $J_{r}(\pi) \triangleq E^{\pi}[\sum_{t=0}^{\infty}\gamma^{t}r(s_{t},a_{t})|s_{0}\sim P_{0}]$, where $P_{0}$ is the initial state distribution. In our setting, when a RL policy $\pi_{A}$ is not optimal, it means that the policy cannot find a series of actions to maximize the cumulative reward: $\pi_{A} \notin argmax_{\pi}J_{r}(\pi)$. 

\subsection{Technical Overview}
\par We design a framework that leverages LLM-based explanations to refine reinforcement learning, as shown in Figure~\ref{fig:model}. Our method consists of three main stages: (i) \textbf{Identification: }given a pre-trained, sub-optimal RL policy, we first collect its historical trajectories. These trajectories are then processed using a state interpretation function that converts the state transitions within each episode into natural language descriptions. We combine these descriptions with environment-specific information and a set of identification instructions to form a prompt for the LLM. The LLM uses this prompt to identify the critical states—those most influential to future rewards—in each episode.

\textbf{(ii) Improvement. }After identifying the critical states, we involve the LLM in the policy refinement process by engaging it with three components of the Markov Decision Process (MDP):

(a) Action suggestion: For each identified critical state, we prompt the LLM to analyze the environment dynamics and infer the appropriate actions the agent should have taken. These state-action pairs are stored in a lookup table, which the RL agent consults during policy refinement.
(b) Reward generation: The LLM further analyzes the agent’s effective and sub-optimal actions at the identified critical states. Based on this case-by-case analysis, it provides contextual reasoning that informs reward assignment, enabling more targeted and interpretable policy updates.
(c) We combine both (a) and (b), allowing the LLM to contribute to multiple components of the MDP—namely, action selection and reward shaping—during policy refinement.

\subsection{LLM-based Critical State Identification}

\par To refine a reinforcement learning (RL) policy starting from a sub-optimal baseline, it is crucial to identify moments in the agent’s past trajectories where its actions significantly impact the cumulative reward. These moments, or critical states, serve as key targets for subsequent analysis and policy refinement.
While prior work~\cite{guo2021edge, cheng2023statemask, liuutility} has designed methods for identifying critical states from historical trajectories, these approaches typically involve training an auxiliary network. This introduces uncertainty into the refinement process, and training such networks can be computationally expensive, especially in complex environments.
To address this, we leverage the reasoning and in-context learning capabilities of large language models (LLMs) to identify critical states directly from historical trajectories, bypassing the need for additional network training.

Given a set of $N$ historical episodes $H = \{H_{1}, H_{2},..., H_{N}\}$, where $H_{i}=\{(s_{0}, a_{0},r_{0}),...,(s_{T_{i}}, a_{T_{i}},r_{T_{i}})\}, {\forall} i\in N$, we first apply a state interpretation function to convert each numerical trajectory into a natural language description of the state transitions. This translation helps the LLM understand the structure of the state and action spaces, as well as the environmental dynamics influenced by the agent’s behavior. 

\begin{figure}[h]
    \centering
    \includegraphics[width=1\linewidth, keepaspectratio=true]{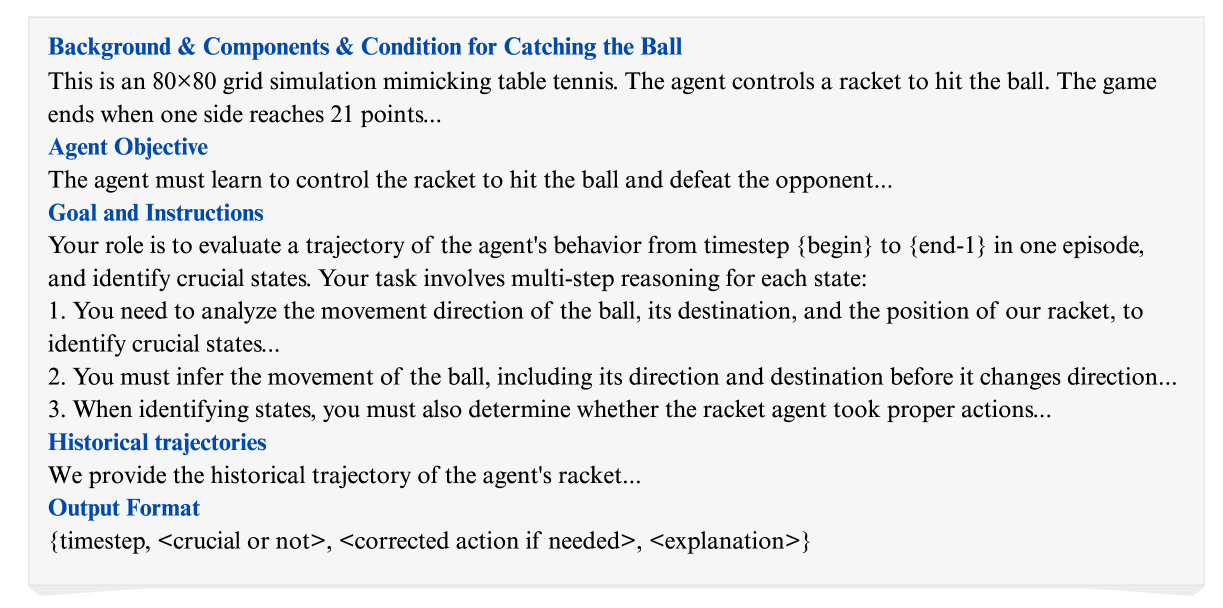}
    \vspace{-15pt}
    \caption{A simplified version of the prompt for identifying critical states in the Pong environment}
    \label{fig:prompt}
    \vspace{-10pt}
\end{figure}

To guide the LLM in identifying critical states, we construct a structured prompt consisting of the following components: (1) relevant environment information such as task goals and termination conditions, (2) multi-step reasoning instructions, (3) a natural language description of the agent’s past trajectory, and (4) output format. The structure of this prompt is illustrated in Figure~\ref{fig:prompt}.
For specific environments, we augment the prompt with environment-specific criteria. For example, in the Pong environment, we inform the LLM that an episode terminates when the ball’s y-coordinate exceeds a threshold (e.g., 70), and instruct it to treat states as critical if actions taken at those points are likely to lead to early termination.
To support accurate reasoning, the prompt first asks the LLM to estimate the agent’s movement and interaction patterns. We explicitly define a critical state as one in which the agent’s action has a significant influence on its ability to maximize future rewards. Additionally, we require the LLM to provide a rationale for each identified critical state, which enhances interpretability and supports downstream policy refinement.
Further details about the prompt design are provided in \hyperref[appendix:B]{Appendix~B}.

\subsection{LLM Explanation-based RL Policy Refinement}
\par After LLM identifies critical states in the agent's historical trajectories, we let LLM get involved in different components of the MDP to support policy refinement. 

\par \textbf{Action Suggestion:} To help the agent select more optimal actions in critical states and thereby improve cumulative reward, we use the LLM to suggest better alternatives to the agent’s original decisions. This approach is inspired by prior work~\cite{li2022pre, ahn2022can} that demonstrates the effectiveness of LLMs as decision-makers in generating candidate actions to guide RL agents.
Specifically, we use a prompt that incorporates the LLM-identified critical states and corresponding explanations to instruct the LLM to recommend actions for those states.  

The resulting state-action pairs—each consisting of a critical state and the corresponding LLM-suggested action—are stored in a lookup table. During policy refinement, the agent checks whether the current state $s_t$ appears in the lookup table. If the state is not critical, the agent proceeds with its original policy action $a_t$; otherwise, it executes the LLM-suggested action $a_t^{LLM}$. The final action taken by the agent, denoted as $a_t^{true}$, is defined as:
\vspace{-5pt}
\begin{equation}
a_{t}^{true} = 
\begin{cases}
    a_{t}^{LLM}, & \text{if } s_{t} \text{ is critical} \\
    a_{t},  & \text{if } s_{t} \text{ is not critical}
\end{cases}
\end{equation}
\vspace{-5pt}

\par After the true action is determined, the agent will execute it in the environment. Through the above process, we utilize LLM to get involved in the agent's action choices and correct the agent's actions at critical states in the refinement process to improve its performance.

\par
\begin{figure}[h]
    \centering
    \includegraphics[width=\linewidth, keepaspectratio=true]{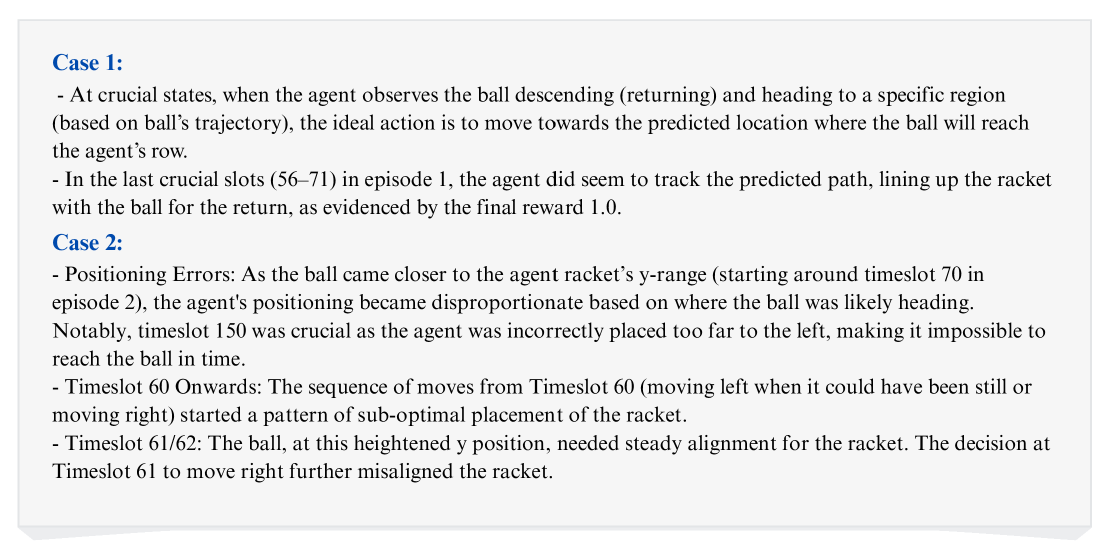}
    \vspace{-15pt}
    \caption{An example of case analysis}
    \label{fig:failure}
    % \vspace{-10pt}
\end{figure}

 \textbf{Reward generation:} We also involve the LLM in the reward generation process to further support policy refinement. To enable the LLM to generate implicit rewards for a non-optimal RL policy, we first prompt it to perform a case analysis for each episode. Specifically, the LLM is asked to analyze and summarize both effective and suboptimal behaviors demonstrated by the agent. To achieve this, we use the critical state identification results as input and construct prompts that guide the LLM to apply its in-context learning and reflection capabilities~\cite{lan2024traj, wang2025steca}. The LLM examines the agent’s actions at each critical state and identifies patterns of success and failure.
Figure~\ref{fig:failure} shows two examples of case analyses in the Pong environment. The LLM is able to diagnose specific failures, such as positioning errors and inaccurate movements, while also recognizing and highlighting effective behaviors. These analyses serve as valuable references for shaping rewards in a way that emphasizes desirable actions and discourages poor ones.

\par After we get the case analysis for the agent's historical trajectories, we incorporate it into a prompt that instructs the LLM to evaluate the agent’s actions and assign a numerical reward in critical states. Given an input state $s_{t}$ and the agent's true action $a_{t}^{true}$, the LLM-generated reward $r_{t}^{LLM}$ is combined with the environment reward $r_{t}^{Env}$ to guide policy learning. The final reward $r_{t}$ is formulated as follows:
\begin{equation}
r_{t} = 
\begin{cases}
    r_{t}^{Env}(s_{t},a_{t}^{true}) + \alpha * r_{t}^{LLM}(s_{t},a_{t}^{true}), & \text{if } s_{t} \text{ is critical} \\
    r_{t}^{Env}(s_{t},a_{t}^{true}),  & \text{if } s_{t} \text{ is not critical}
\end{cases}
\end{equation}

\renewcommand{\algorithmicrequire}{\textbf{Input:}}
\renewcommand{\algorithmicensure}{\textbf{Output:}}
{\renewcommand{\baselinestretch}{1.2}\normalsize

\begin{algorithm}
\small
    \caption{LLM-guided RL Policy Refining}
    \label{ab1}
    \begin{algorithmic}
    \REQUIRE Pre-trained policy $\pi$, initial state distribution $P_{0}$, memory buffer $\mathcal{D}$, lookup table $\mathcal{T}$
    \ENSURE The agent's refined policy $\pi'$
    \FOR{$n = 1$ to $N$}
    \STATE $\mathcal{D} \leftarrow \emptyset$, $s_{0} \sim P_{0}$
    \FOR{$t=0$ to $T$}
    \STATE Sample $a_{t} \sim \pi(a_{t}|s_{t})$
    \STATE \textbf{/* Check whether the state belongs to the look-up table */}
    \IF{$s_{t} \in L $}
    \STATE \textbf{/* Replace the agent's action with the LLM-guided suggested action, and generate a LLM-guided reward*/}
    \STATE Query the corrected action $a_{t}^{LLM}$ in $\mathcal{T}$
    \STATE $a_{t}^{true} = a_{t}^{LLM}$, $r_{t}^{LLM} \leftarrow LLM(s_{t},a_{t}^{true})$ 
    \ELSE
    \STATE $a_{t}^{true} = a_{t}$, $r_{t}^{LLM} \leftarrow 0$ 
    \ENDIF
    \STATE $(s_{t+1},r_{t}^{Env}) \leftarrow env.step(a^{true}_{t})$
    \STATE $r_{t} = r_{t}^{Env} + \alpha * r^{LLM}_{t}$ 
    \STATE Add $(s_{t},a_{t}^{true},s_{t+1},r_{t})$ to $\mathcal{D}$
    \ENDFOR
    \STATE Optimize $\pi_{\theta}$ w.r.t PPO loss on $\mathcal{D}$
    \ENDFOR
    \STATE $\pi' \leftarrow \pi_{\theta}$
    \end{algorithmic}
\end{algorithm}

}

\par where $\alpha$ is the coefficient of the transformation to the environment reward. Given the above designs, we develop three potentially available directions to refine the RL policy: (1) utilize LLM-generated corrected actions to replace the agent's action in critical states; (2) utilize LLM-generated reward combined with environment reward in critical states; (3) utilize the settings of (1) and (2). The detailed steps are in Algorithm~\ref{ab1}.

\section{Evaluation}
\par In this section, we conduct experiments to evaluate our models, answering the following questions:
\vspace{-5pt}
\begin{itemize}[leftmargin=*]
\item RQ1: How does our model perform compared to existing baselines?
\item RQ2: How effective are the individual components of our model in improving RL performance?
\item RQ3: How does the LLM generate explanations for its action suggestions and reward assignments?
\end{itemize}
\vspace{-10pt}
\subsection{Experiment Setup}
\par \textbf{Environment Selection.} 
We evaluate our method on four representative encironments to demonstrate its effectiveness. These include one simple normal-form game—Pong~\cite{brockman2016openai}—and three complex continuous control tasks from the MuJoCo suite: Hopper, Walker2d, and Ant~\cite{todorov2012mujoco}. Detailed settings for each environment are provided in \hyperref[appendix:A]{Appendix~A}.

\par \textbf{Baselines.} We compare three variants of our method:
\textbf{ULTRA-A:} Uses LLM-generated action corrections at critical states.
\textbf{ULTRA-R:} Uses LLM-generated implicit rewards at critical states.
\textbf{ULTRA-RA:} Combines both LLM-generated actions and rewards. 
We also compare against three baseline methods: (i) \textbf{RICE:}~(\cite{cheng2024rice}) It is a state-of-the-art RL policy refinement method which trains a mask network first to identify the agent's critical states and fine-tunes the agent's policy starting at initial states and those critical states. (ii) \textbf{LIR:}~(\cite{zheng2018learning}) It is a method that aims to learn an intrinsic reward to formulate the shaping reward function of the environment. (iii) \textbf{HLC:}~\cite{savinov2018episodic,goyal2019using} We introduce human-in-the-loop correction that provides a reward function designed by humans to guide policy learning. \cite{goyal2019using} designs a natural language-based reward function for the Pong game, and \cite{savinov2018episodic} designs a space-based reward function for the Ant game. 
All methods are built on top of PPO~\cite{schulman2017proximal}, which serves as the base RL algorithm.

\par \textbf{Evaluation Metrics:} We evaluate policy refinement performance by measuring the average cumulative return across 100 episodes. Results are reported with the mean and standard deviation across multiple random seeds.

\par \textbf{Implementation:} We implement our method and baselines using PyTorch 1.9.1 in a Python 3.8 environment and train them on four NVIDIA RTX A5000 GPUs, each with 24 GB of memory. For LLMs, we choose GPT-4o as our LLM to identify the critical states in the agent's historical trajectories, correct the agent's actions at those critical states, and provide a reward to guide the agent's refinement. By testing the performance under different hyperparameter settings, we use the following settings: the coefficient of the transformation of LLM-generated reward to the environment reward $\alpha$ is best at 0.5 for the Pong experiment and 0.1 for the MuJoCo experiments, tested among [0.1, 0.5, 1.0]. The learning rate is best at 1e-4, tested among [1e-3, 1e-4, 1e-5]. The batch size is 64. Other detailed experiment settings can be seen in \hyperref[appendix:A]{Appendix~A}.

\subsection{Overall Performance (RQ1)}

\textbf{Simple Sparse-reward Game.} To show the effectiveness of our model in the sparse-reward environment, we conduct the comparison experiments in the Pong environment, which is an 80×80 grid simulation game that mimics table tennis. In this environment, the agent only receives non-zero rewards when the game is won (+1) or lost (–1); all other timesteps yield zero reward. As shown in Table~\ref{tab:performance}, all three variants of our method outperform the baselines, with ULTRA-RA achieving the most significant improvement over the base model. Several factors may contribute to this performance gain: (1) compared to the intrinsic reward-based model (e.g., LIR), the LLM-generated rewards and corrected actions provide more informative and efficient guidance for policy refinement. (2) In sparse-reward environments, leveraging the LLM’s in-context learning and reflection capabilities to analyze complete trajectories allows for more accurate identification of critical states. This appears more effective than reward-difference–based methods like RICE, which rely solely on observed reward changes to detect important moments. (3) Human-designed reward functions applied at key timesteps (e.g., assigning positive reward when the paddle hits the ball) lack sufficient granularity to meaningfully shape behavior, limiting their effectiveness for policy refinement. (4) The comparison between ULTRA-R and ULTRA-A indicates that, in sparse-reward settings, providing informative rewards at critical states is more beneficial than correcting actions alone.
Furthermore, the superior performance of ULTRA-RA over both ULTRA-R and ULTRA-A suggests that the two refinement strategies—LLM-guided action correction and reward generation—are complementary rather than conflicting. To validate this, we also prompt the LLM to generate refinement strategies based on its case analyses. As shown in \hyperref[appendix:C]{Appendix~C}, the LLM’s generated strategies are consistent with our design, further supporting the effectiveness of our approach.

\begin{table}[t]
\centering
\caption{Performance Comparison of Different Approaches. Bolded scores indicate the best. The average improvement of ULTRA-RA (our best-performing model) over three baselines is shown at the bottom row.}
\setlength\tabcolsep{12pt}
\label{tab:performance}
\resizebox{0.9\linewidth}{!}{
\begin{tabular}{@{}ccccc@{}}
\toprule
\textbf{Method} & \textbf{Pong} & \textbf{Hopper} & \textbf{Walker2d} & \textbf{Ant} \\
\midrule
PPO & 0.3 ($\pm$0.2) & 3571.68 ($\pm$85.78) & 3639.96 ($\pm$98.73) & 5865.47 ($\pm$127.26) \\
RICE & 0.3 ($\pm$0.2) & 3756.24 ($\pm$69.47) & 3721.58 ($\pm$55.96) & 5932.36 ($\pm$91.81) \\
LIR & 0.35 ($\pm$0.2) & 3778.62 ($\pm$55.58) & 3729.74 ($\pm$54.78) & 5974.65 ($\pm$91.37) \\
HLC & 0.3 ($\pm$0.2) & - & - & 5912.03 ($\pm$89.96) \\
\midrule
ULTRA-A & 0.4 ($\pm$0.2) & 3842.43 ($\pm$70.39) & 3804.27 ($\pm$76.82) & 6024.39 ($\pm$94.05) \\
ULTRA-R & 0.5 ($\pm$0.1) & 3890.54 ($\pm$70.94) & 3957.03 ($\pm$73.46) & 6102.4 ($\pm$95.83) \\
ULTRA-RA & \textbf{0.8 ($\pm$0.1)} & \textbf{3986.52 ($\pm$71.72)} & \textbf{4206.15 ($\pm$84.17)} & \textbf{6249.76 ($\pm$96.33)} \\
\midrule
Improvement & 165$\%$ & 7.68$\%$ & 13.77$\%$ & 5.55$\%$ \\ 
\bottomrule
\end{tabular}
}
\vspace{-10pt}
\end{table}

\textbf{Complex Dense-reward Game.} 
We further evaluate our model in complex, dense-reward environments—specifically, three MuJoCo tasks: Hopper, Walker2d, and Ant. In these environments, the agent must learn to control a robotic body to move forward quickly while maintaining balance and avoiding large pitch angles. Rewards are provided continuously, reflecting forward progress, survival, and stability.
As shown in Table~\ref{tab:performance}, all three versions of our model outperform the baselines, with ULTRA-RA achieving the highest performance across all three tasks. Several factors may explain these results:
(1) Although dense-reward settings provide more frequent feedback, allowing RICE to detect reward changes when masking actions, the complexity of the control tasks requires the agent to take precise and coordinated actions at nearly every timestep. This makes it difficult for RICE to reliably identify critical states through action masking alone. (2) Intrinsic reward-based methods (e.g., LIR) may offer limited benefit in these settings, where the environment already provides rich and informative rewards. (3) Human-designed shaped reward functions applied at select timesteps (e.g., assigning positive reward only when the agent exceeds a certain distance) may fail to capture the continuous, fine-grained adjustments required for effective control, limiting their impact on policy refinement. (4) Although ULTRA-R still outperforms ULTRA-A, the relative advantage of LLM-generated rewards over LLM-suggested actions is less pronounced in dense-reward environments. This may be due to the fact that dense environment rewards already provide strong supervision, which can dilute the influence of additional reward signals generated by the LLM. In addition, the consistently strong performance of ULTRA-RA across all tasks confirms that LLM-guided action correction and reward generation remain complementary. Even in the complex dense-reward settings, they can jointly support effective and non-conflicting policy refinement.

\subsection{Abaltion Study (RQ2)}

\par \textbf{The significance of LLM-based critical state identification. }To evaluate the importance of LLM-based critical state identification in policy refinement, we introduce a variant of our model, MASK-R, in which the critical state identification module is replaced with the StateMask model~\cite{cheng2023statemask}. The critical states identified by StateMask are then passed to a prompt that instructs the LLM to generate implicit rewards to guide the agent’s refinement.
As shown in Table~\ref{tab:abs}, MASK-R outperforms RICE but underperforms ULTRA-R. This result highlights two key observations:
(1) Critical states identified by the LLM are more effective in supporting another LLM’s reasoning when generating instructive rewards, compared to those identified by StateMask. This underscores the importance of LLM-based identification for downstream refinement.
(2) LLM-generated rewards provide more targeted and meaningful guidance for policy improvement than exploration-driven intrinsic rewards, further validating the contribution of LLM explanations in the refinement process.

\begin{table}[t]
\setlength\tabcolsep{6pt}
\centering
\caption{Performance Comparison of Different Approaches}
\label{tab:abs}
\resizebox{0.8\linewidth}{!}{
\begin{tabular}{@{}ccccc@{}}
\toprule
\textbf{Scenario} & \textbf{RICE} & \textbf{MASK-R} & \textbf{LLM-RND} & \textbf{ULTRA-R} \\
\midrule
Pong  & 0.3 ($\pm$0.2) & 0.4 ($\pm$0.1) & 0.35 ($\pm$0.2) & 0.5 ($\pm$0.1) \\
Hopper & 3756.24 ($\pm$69.47) & 3801.56 ($\pm$69.39) & 3788.89 ($\pm$69.86) & 3890.54 ($\pm$70.94) \\
\bottomrule
\end{tabular}
}
\vspace{-10pt}
\end{table}

\par \textbf{The significance of LLM-based reward assignment. }To assess the contribution of LLM-based reward assignment, we introduce a variant of our model, LLM-RND, in which the reward generation module is replaced with a novelty-driven exploration mechanism, RND~\cite{burda2018exploration}, to guide policy refinement.
As shown in Table~\ref{tab:abs}, LLM-RND outperforms RICE but underperforms ULTRA-R. These results suggest the following:
Although the critical states identified by the LLM are more informative than those identified by StateMask (as discussed previously), the novelty-driven intrinsic rewards provided by RND are less effective for policy refinement. In contrast, LLM-generated rewards—derived from a case-based analysis of the agent’s state-action trajectories—offer more meaningful and targeted feedback. This contextual reasoning enables the LLM to assign instructive rewards that better guide the agent’s learning process, highlighting the importance of explanation-driven reward assignment in our framework.

\begin{figure}[t]
    \centering
    \includegraphics[width=1\linewidth, keepaspectratio=true]{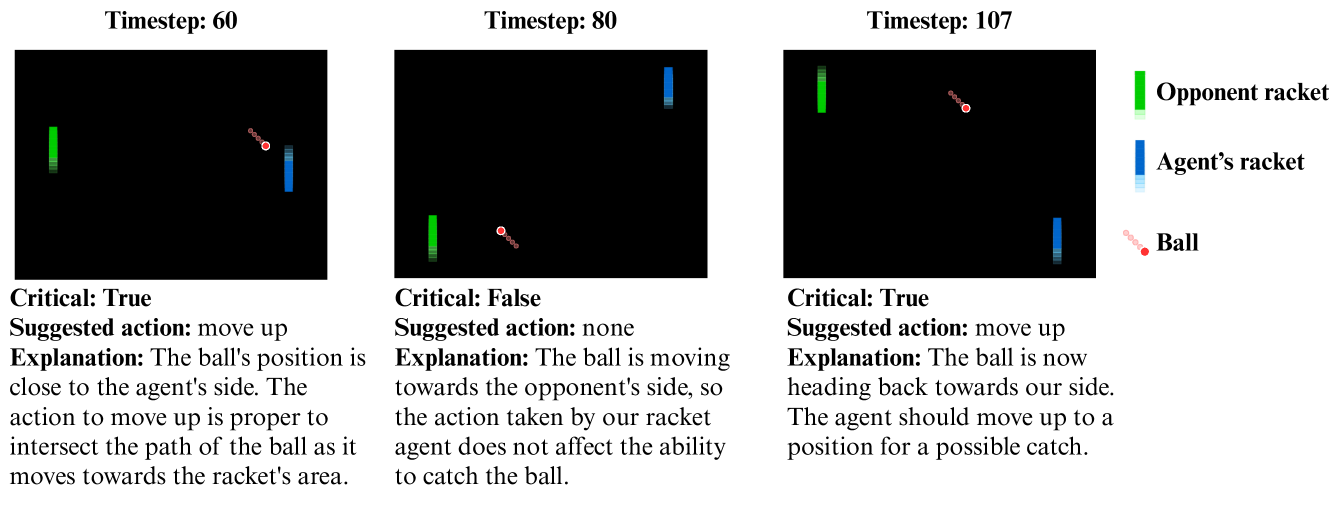}
    \vspace{-10pt}
    \caption{The identifications and action suggestions in three timesteps}
    \label{fig:case study}
    \vspace{-10pt}
\end{figure}

\begin{figure}[t]
    \centering
    \includegraphics[width=1\linewidth, keepaspectratio=true]{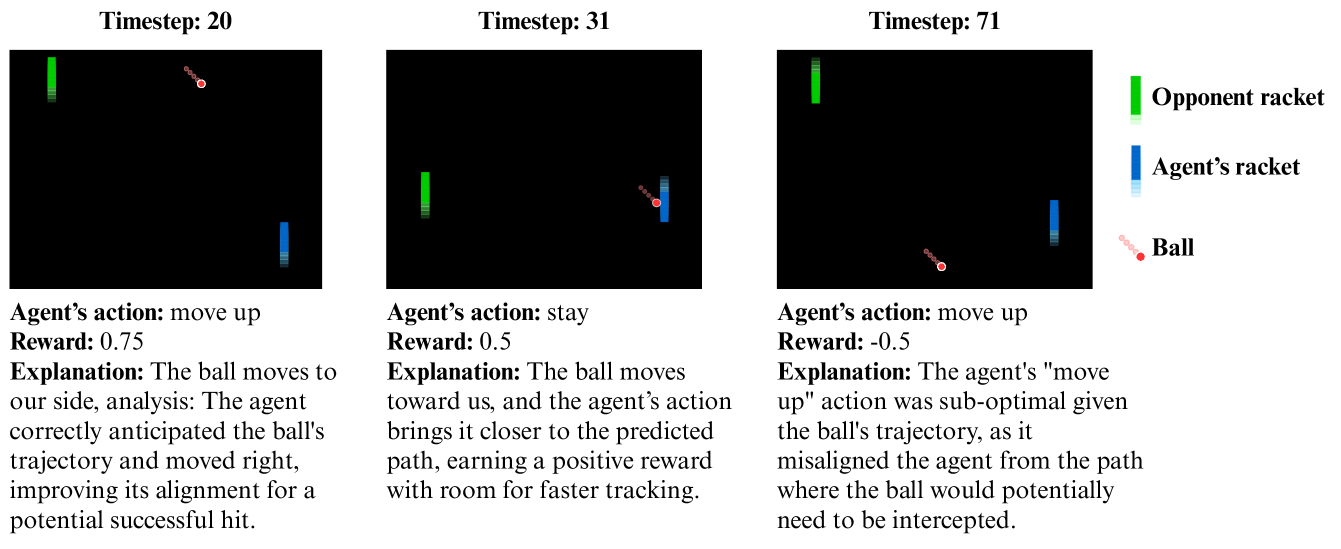}
    \vspace{-10pt}
    \caption{The LLM-generated rewards in three timesteps}
    \label{fig:reward_case}
    \vspace{-10pt}
\end{figure}

\subsection{Case Study (RQ3)}
\par We conduct a case study using the Pong environment to illustrate the capabilities of our model in identifying critical states, generating action suggestions, and assigning rational rewards. Figure~\ref{fig:case study} presents three timesteps from a single episode, highlighting the LLM-based critical state identification along with the corresponding action suggestions. The figure demonstrates that the LLM is able to infer the ball’s movement direction across continuous states and estimate its velocity. This enables the model to assess whether a given state poses a threat to the paddle’s ability to intercept the ball—an essential factor in determining criticality.
Figure~\ref{fig:reward_case} shows the LLM-generated rewards for the same three timesteps. The results reveal that the LLM can not only assign positive rewards to reinforce correct actions but also penalize sub-optimal decisions, demonstrating its effectiveness in guiding RL policy refinement.
Moreover, the reasoning behind LLM-identified critical states, suggested actions, and reward assignments enhances the interpretability of the refinement process. Full examples of the LLM-generated explanations are provided in \hyperref[appendix:C]{Appendix~C}.

\section{Conclusion}
\par We introduce ULTRA, a framework that leverages large language models (LLMs) to address training bottlenecks in reinforcement learning (RL). Specifically, we first prompt the LLM to identify critical states from the historical trajectories of a pre-trained, sub-optimal agent. These identified states are then used as the basis for further prompts, instructing the LLM to generate suggested actions and implicit rewards to guide policy refinement.
We explore three complementary strategies for integrating LLM-generated insights into the RL training process. Experimental results across multiple environments show that our approach consistently outperforms state-of-the-art baselines, demonstrating the effectiveness of LLM-based explanations in enhancing RL policy refinement.

\bibliographystyle{plain}
\bibliography{refs}

\begin{thebibliography}{10}

\bibitem{agarwal2022reincarnating}
Rishabh Agarwal, Max Schwarzer, Pablo~Samuel Castro, Aaron~C Courville, and Marc Bellemare.
\newblock Reincarnating reinforcement learning: Reusing prior computation to accelerate progress.
\newblock {\em Advances in neural information processing systems}, 35:28955--28971, 2022.

\bibitem{ahn2022can}
Michael Ahn, Anthony Brohan, Noah Brown, Yevgen Chebotar, Omar Cortes, Byron David, Chelsea Finn, Chuyuan Fu, Keerthana Gopalakrishnan, Karol Hausman, et~al.
\newblock Do as i can, not as i say: Grounding language in robotic affordances.
\newblock {\em arXiv preprint arXiv:2204.01691}, 2022.

\bibitem{brockman2016openai}
Greg Brockman, Vicki Cheung, Ludwig Pettersson, Jonas Schneider, John Schulman, Jie Tang, and Wojciech Zaremba.
\newblock Openai gym.
\newblock {\em arXiv preprint arXiv:1606.01540}, 2016.

\bibitem{burda2018exploration}
Yuri Burda, Harrison Edwards, Amos Storkey, and Oleg Klimov.
\newblock Exploration by random network distillation.
\newblock {\em arXiv preprint arXiv:1810.12894}, 2018.

\bibitem{cai2023imitation}
Xin-Qiang Cai, Yu-Jie Zhang, Chao-Kai Chiang, and Masashi Sugiyama.
\newblock Imitation learning from vague feedback.
\newblock {\em Advances in Neural Information Processing Systems}, 36:48275--48292, 2023.

\bibitem{cheng2023statemask}
Zelei Cheng, Xian Wu, Jiahao Yu, Wenhai Sun, Wenbo Guo, and Xinyu Xing.
\newblock Statemask: Explaining deep reinforcement learning through state mask.
\newblock {\em Advances in Neural Information Processing Systems}, 36:62457--62487, 2023.

\bibitem{cheng2024rice}
Zelei Cheng, Xian Wu, Jiahao Yu, Sabrina Yang, Gang Wang, and Xinyu Xing.
\newblock Rice: Breaking through the training bottlenecks of reinforcement learning with explanation.
\newblock {\em arXiv preprint arXiv:2405.03064}, 2024.

\bibitem{goyal2019using}
Prasoon Goyal, Scott Niekum, and Raymond~J Mooney.
\newblock Using natural language for reward shaping in reinforcement learning.
\newblock {\em arXiv preprint arXiv:1903.02020}, 2019.

\bibitem{guan2021widening}
Lin Guan, Mudit Verma, Suna~Sihang Guo, Ruohan Zhang, and Subbarao Kambhampati.
\newblock Widening the pipeline in human-guided reinforcement learning with explanation and context-aware data augmentation.
\newblock {\em Advances in Neural Information Processing Systems}, 34:21885--21897, 2021.

\bibitem{guo2021edge}
Wenbo Guo, Xian Wu, Usmann Khan, and Xinyu Xing.
\newblock Edge: Explaining deep reinforcement learning policies.
\newblock {\em Advances in Neural Information Processing Systems}, 34:12222--12236, 2021.

\bibitem{huang2024uncertainty}
Wenhui Huang, Zitong Shan, Shanhe Lou, and Chen Lv.
\newblock Uncertainty-aware reinforcement learning for autonomous driving with multimodal digital driver guidance.
\newblock In {\em 2024 IEEE International Conference on Robotics and Automation (ICRA)}, pages 18355--18361. IEEE, 2024.

\bibitem{huang2022inner}
Wenlong Huang, Fei Xia, Ted Xiao, Harris Chan, Jacky Liang, Pete Florence, Andy Zeng, Jonathan Tompson, Igor Mordatch, Yevgen Chebotar, et~al.
\newblock Inner monologue: Embodied reasoning through planning with language models.
\newblock {\em arXiv preprint arXiv:2207.05608}, 2022.

\bibitem{kwon2023reward}
Minae Kwon, Sang~Michael Xie, Kalesha Bullard, and Dorsa Sadigh.
\newblock Reward design with language models.
\newblock {\em arXiv preprint arXiv:2303.00001}, 2023.

\bibitem{laleh2024survey}
Alireza~Rashidi Laleh and Majid~Nili Ahmadabadi.
\newblock A survey on enhancing reinforcement learning in complex environments: Insights from human and llm feedback.
\newblock {\em arXiv preprint arXiv:2411.13410}, 2024.

\bibitem{lan2024traj}
Zhengxing Lan, Lingshan Liu, Bo~Fan, Yisheng Lv, Yilong Ren, and Zhiyong Cui.
\newblock Traj-llm: A new exploration for empowering trajectory prediction with pre-trained large language models.
\newblock {\em IEEE Transactions on Intelligent Vehicles}, 2024.

\bibitem{li2022pre}
Shuang Li, Xavier Puig, Chris Paxton, Yilun Du, Clinton Wang, Linxi Fan, Tao Chen, De-An Huang, Ekin Aky{\"u}rek, Anima Anandkumar, et~al.
\newblock Pre-trained language models for interactive decision-making.
\newblock {\em Advances in Neural Information Processing Systems}, 35:31199--31212, 2022.

\bibitem{liuutility}
Shicheng Liu and Minghui Zhu.
\newblock Utility: Utilizing explainable reinforcement learning to improve reinforcement learning.
\newblock In {\em The Thirteenth International Conference on Learning Representations}, 2025.

\bibitem{metz2024mapping}
Yannick Metz, David Lindner, Rapha{\"e}l Baur, and Mennatallah El-Assady.
\newblock Mapping out the space of human feedback for reinforcement learning: A conceptual framework.
\newblock {\em arXiv preprint arXiv:2411.11761}, 2024.

\bibitem{savinov2018episodic}
Nikolay Savinov, Anton Raichuk, Rapha{\"e}l Marinier, Damien Vincent, Marc Pollefeys, Timothy Lillicrap, and Sylvain Gelly.
\newblock Episodic curiosity through reachability.
\newblock {\em arXiv preprint arXiv:1810.02274}, 2018.

\bibitem{schulman2017proximal}
John Schulman, Filip Wolski, Prafulla Dhariwal, Alec Radford, and Oleg Klimov.
\newblock Proximal policy optimization algorithms.
\newblock {\em arXiv preprint arXiv:1707.06347}, 2017.

\bibitem{shridhar2023perceiver}
Mohit Shridhar, Lucas Manuelli, and Dieter Fox.
\newblock Perceiver-actor: A multi-task transformer for robotic manipulation.
\newblock In {\em Conference on Robot Learning}, pages 785--799. PMLR, 2023.

\bibitem{tan2023joint}
Heng Tan, Yukun Yuan, Shuxin Zhong, and Yu~Yang.
\newblock Joint rebalancing and charging for shared electric micromobility vehicles with energy-informed demand.
\newblock In {\em Proceedings of the 32nd ACM International Conference on Information and Knowledge Management}, pages 2392--2401, 2023.

\bibitem{todorov2012mujoco}
Emanuel Todorov, Tom Erez, and Yuval Tassa.
\newblock Mujoco: A physics engine for model-based control.
\newblock In {\em 2012 IEEE/RSJ international conference on intelligent robots and systems}, pages 5026--5033. IEEE, 2012.

\bibitem{van2022correct}
Sanne Van~Waveren, Christian Pek, Jana Tumova, and Iolanda Leite.
\newblock Correct me if i'm wrong: Using non-experts to repair reinforcement learning policies.
\newblock In {\em 2022 17th ACM/IEEE International Conference on Human-Robot Interaction (HRI)}, pages 493--501. IEEE, 2022.

\bibitem{vinyals2019grandmaster}
Oriol Vinyals, Igor Babuschkin, Wojciech~M Czarnecki, Micha{\"e}l Mathieu, Andrew Dudzik, Junyoung Chung, David~H Choi, Richard Powell, Timo Ewalds, Petko Georgiev, et~al.
\newblock Grandmaster level in starcraft ii using multi-agent reinforcement learning.
\newblock {\em nature}, 575(7782):350--354, 2019.

\bibitem{wang2025steca}
Hanlin Wang, Jian Wang, Chak~Tou Leong, and Wenjie Li.
\newblock Steca: Step-level trajectory calibration for llm agent learning.
\newblock {\em arXiv preprint arXiv:2502.14276}, 2025.

\bibitem{wu2023read}
Yue Wu, Yewen Fan, Paul~Pu Liang, Amos Azaria, Yuanzhi Li, and Tom~M Mitchell.
\newblock Read and reap the rewards: Learning to play atari with the help of instruction manuals.
\newblock {\em Advances in Neural Information Processing Systems}, 36:1009--1023, 2023.

\bibitem{yao2020keep}
Shunyu Yao, Rohan Rao, Matthew Hausknecht, and Karthik Narasimhan.
\newblock Keep calm and explore: Language models for action generation in text-based games.
\newblock {\em arXiv preprint arXiv:2010.02903}, 2020.

\bibitem{yin2024offline}
Zhao-Heng Yin and Pieter Abbeel.
\newblock Offline imitation learning through graph search and retrieval.
\newblock {\em arXiv preprint arXiv:2407.15403}, 2024.

\bibitem{yu2023airs}
Jiahao Yu, Wenbo Guo, Qi~Qin, Gang Wang, Ting Wang, and Xinyu Xing.
\newblock $\{$AIRS$\}$: Explanation for deep reinforcement learning based security applications.
\newblock In {\em 32nd USENIX Security Symposium (USENIX Security 23)}, pages 7375--7392, 2023.

\bibitem{yu2023language}
Wenhao Yu, Nimrod Gileadi, Chuyuan Fu, Sean Kirmani, Kuang-Huei Lee, Montse~Gonzalez Arenas, Hao-Tien~Lewis Chiang, Tom Erez, Leonard Hasenclever, Jan Humplik, et~al.
\newblock Language to rewards for robotic skill synthesis.
\newblock {\em arXiv preprint arXiv:2306.08647}, 2023.

\bibitem{zheng2018learning}
Zeyu Zheng, Junhyuk Oh, and Satinder Singh.
\newblock On learning intrinsic rewards for policy gradient methods.
\newblock {\em Advances in neural information processing systems}, 31, 2018.

\end{thebibliography}

\section*{A. Details of Evaluation}
\label{appendix:A}
\subsection*{A1. Baseline Implementation.} Regarding the baseline RICE~\cite{cheng2024rice}, which combines the critical states and initial states as a new initial state distribution, and adopts an exploration-encouraged method to refine the base model. To achieve the replay function, in the Pong game, we utilize $env.restore\_full\_state$ to reset the environment state to a certain critical state. In the MuJoCo environment, we utilize $mujoco.mj\_forward$ to write a $restore\_full\_state$ function for the MuJoCo environment. For its detailed implementation, we use their released open-sourced code from \url{https://github.com/chengzelei/RICE}. Regarding the implementation of LIR~\cite{zheng2018learning}, we utilize the open-sourced code from \url{https://github.com/Hwhitetooth/lirpg}.

\subsection*{A2. Game Introduction}
\begin{itemize}[leftmargin=*]
\item \textbf{Pong:} It is an 80×80 grid simulation game that mimics table tennis (ping-pong): control the racket to hit the ball sent by the opponent. Whoever fails to catch it loses (similar to table tennis). The winner’s reward is 1 point, and the loser’s reward is -1 point. The game ends when the first person reaches 21 points. We assign our racket with an agent to learn to control how to move in order to hit the ball sent by the opponent and defeat the opponent (first to reach 21 points).
\item \textbf{Hopper:} It is a continuous control environment based on the MuJoCo physics engine, used for benchmarking reinforcement learning algorithms. It simulates a planar robot (essentially a one-legged creature) trying to hop forward without falling. The game episode ends if any of the following occur: (1) Falling: torso vertical position falls below 0.7 meters, (2) Large Pitch Angle: Absolute torso pitch angle exceeds 36 degrees ($abs$(pitch) > 0.628 radians). We assign this robot with an agent to make the Hopper hop forward as far as possible while maintaining balance.
\item \textbf{Walker-2d:} It is a continuous-control bipedal robot environment based on the MuJoCo physics engine, used for benchmarking reinforcement learning algorithms. It simulates a two-legged robot (similar to a simplified bipedal walker) whose objective is to move forward as quickly and stably as possible without falling. The episode game ends if any of the following occur: (1) Falling: vertical position of the torso falls below 0.8 meters or exceeds 2.0 meters, (2) Invalid Pitch Angle: pitch angle of the torso exceeds around $\pm$1 radian (approximately $\pm$57 degrees).
\item \textbf{Ant:} It is a continuous-control robot environment based on the MuJoCo physics engine, used for benchmarking reinforcement learning algorithms. It simulates the ant robot moving forward (maximizing the positive x-distance) while trying not to fall over or make unnecessary large movements (to reduce energy consumption). The episode game ends if the agent's torso height falls below a certain threshold, typically z $\leq$ 0.2, where z is the vertical position of the torso.
\end{itemize}

\section*{B. Prompt Designs}
\label{appendix:B}
\par Here we list the full prompts used in our method.

\par \textbf{(i) Critical State Identification and Action Suggestion in Pong}

\begin{tcolorbox}[mypromptbox, title=Critical State Identification and Action Suggestion in Pong]
\textbf{Background}\par
It is an $80$×$80$ grid simulation game that mimics table tennis (ping-pong): control the racket to hit the ball sent by the opponent. Whoever fails to catch it loses. The winner’s reward is $1$ point, and the loser’s reward $-1$ point. The game ends when the first person reaches $21$ points.

\textbf{Components of this game}\par
Our racket, the opponent racket, and ball move in this $80$×$80$ grid simulation game. Both our racket and the opponent racket are a $2$×$8$ vector, and ball is a $1$×$2$ vector. Both our racket and the opponent racket can only move horizontally, which means that their y coordinates will not change. Therefore, we use the tuple \{minimal x, maximal x, minimal y, maximal y\} to denote the positions of rackets. We use the tuple \{x1, x2, y1, y2\} to denote the position of ball. We use the above tuple to denote the movement range of the opponent racket: \{$0$, 7, 8, 9\} $\sim$ \{73, 80, 8, 9\}; We use the above tuple to denote the movement range of our racket: \{0, 7, 70, 71\} $\sim$ \{73, 80, 70, 71\}.

\textbf{Agent objective}\par
We assign our racket with an agent to learn to control how to move in order to hit the ball sent by the opponent and defeat the opponent (first to reach 21 points). We model the game as MDP: agent observes its state and takes an action based on its policy and updates its policy based on the reward from the environment. The process repeats until the game ends.

\textbf{How to catch the ball}\par
Only if the ball is close to the 8 coordinate points on the upper surface of the racket. Here is an example: if the ball location is \{52, 69, 53, 69\}, only if the location of our racket is from \{52, 59, 70, 71\} to \{46, 53, 70, 71\}, our racket can catch the ball.

\textbf{Identification instructions}\par
(i) This is the historical trajectory in the episode ``\texttt{str(e)}''. You must identify all the critical states in this episode. To understand critical states, here is an example to illustrate what are uncritical states: after the racket agent launches the ball and when the ball is moving towards its opponent, the agent has a few time steps of freedom to choose random actions. This is because the agent’s actions at these steps will not change the movement of the ball and will be less likely to influence the opponent’s actions as well (given the opponent’s focus is on the ball). Therefore, those few time steps are uncritical states for the racket agent.

(ii) You need to use reasoning to determine whether the action in each state has an impact on future rewards, that is, whether it affects the agent's ability to catch and launch the ball.

(iii) You need to analyze the movement direction of the ball, the destination of the ball based on movement direction, and the position of our racket, to infer whether the racket's current state is important, that is, whether the current action affects the racket's ability to approach and successfully catch the ball.

(iv) When you identify if those states are critical or not, you must also infer if the racket agent took proper actions at those critical states. For example, when the ball was moving to our agent's side, our racket agent's actions made it move away from the final destination of the ball. That means that our racket agent made improper actions. You should correct the agent's actions by outputting the corrected actions (\texttt{stay}/\texttt{move left}/\texttt{move right}).

\textbf{Output format}\par
The format of your identified results must be: \{\texttt{timeslot str(begin)}, \texttt{<critical or not>}, \texttt{<corrected actions if agents' actions are unproper>}, \texttt{<explanation>}\}, \{\texttt{timeslot str(begin+1)}, \texttt{<critical or not>}, \texttt{<corrected actions if agents' actions are unproper>}, \texttt{<explanation>}\}, ..., \{\texttt{timeslot str(end-1)}, \texttt{<critical or not>}, \texttt{<corrected actions if agents' actions are unproper>}, \texttt{<explanation>}\}.

You must output the identified results state by state. For corrected actions, you must choose one of three possible actions: \texttt{stay}, \texttt{move left}, or \texttt{move right}. You must give an explanation about why you think those states are critical or not and why you choose that action as the corrected action. If the agent's action is proper at a critical state, you do not have to give a corrected action for that timeslot. If the state is uncritical, the corrected action is \texttt{<none>}.
\end{tcolorbox}

\par \textbf{(ii) Case Analysis in Pong}

\begin{tcolorbox}[mypromptbox, title=Case Analysis in Pong]
\textbf{Background}\par
It is an $80$×$80$ grid simulation game that mimics table tennis (ping-pong): control the racket to hit the ball sent by the opponent. Whoever fails to catch it loses. The winner’s reward is $1$ point, and the loser’s reward $-1$ point. The game ends when the first person reaches $21$ points.

\textbf{Components of this game}\par
Our racket, the opponent's racket, and the ball move in this $80$×$80$ grid simulation game. Both our racket and the opponent's racket are a $2$×$8$ vector, and ball is a $1$×$2$ vector. Both our racket and the opponent's racket can only move horizontally, which means that their y coordinates will not change. Therefore, we use the tuple \{minimal x, maximal x, minimal y, maximal y\} to denote the positions of rackets. We use the tuple \{x1, x2, y1, y2\} to denote the position of ball. We use the above tuple to denote the movement range of the opponent racket: \{$0$, 7, 8, 9\} $\sim$ \{73, 80, 8, 9\}; We use the above tuple to denote the movement range of our racket: \{0, 7, 70, 71\} $\sim$ \{73, 80, 70, 71\}.

\textbf{Agent objective}\par
We assign our racket with an agent to learn to control how to move in order to hit the ball sent by the opponent and defeat the opponent (first to reach 21 points). We model the game as an MDP: the agent observes its state and takes an action based on its policy and updates its policy based on the reward from the environment. The process repeats until the game ends.

\textbf{How to catch the ball}\par
Only if the ball is close to the 8 coordinate points on the upper surface of the racket. Here is an example: if the ball location is \{52, 69, 53, 69\}, only if the location of our racket is from \{52, 59, 70, 71\} to \{46, 53, 70, 71\}, our racket can catch the ball.

\textbf{Analysis instructions}\par
(i) This is the historical trajectory in the episode " + \texttt{str(e)} + ". You must identify all the critical states in this episode. To understand critical states, here is an example to illustrate what is uncricial states: after the racket agent launches the ball and when the ball is moving towards its opponent, the agent has a few time steps of freedom to choose random actions. This is because the agent’s actions at these steps will not change the movement of the ball and will be less likely to influence the opponent’s actions as well (given the opponent’s focus is on the ball). 

(ii) After you identify critical states in the historical trajectory, you must analyze why our racket lost this episode. For example, starting from a certain critical state, our racket took the wrong movements or our racket moved to the ball's destination too late, so that our racket could not catch the ball. Identifying critical states may help you analyze the failure of our racket. 

(iii) You also must analyze when our racket took the great action at those critical states, and how those actions should be rewarded.

\textbf{Output format}\par
The format of your analysis results must be: \{\texttt{timeslot str(begin)}, \texttt{<critical or not>}\}, \{\texttt{timeslot str(begin+1)}, \texttt{<critical or not>}\}, ..., \{\texttt{timeslot str(end-1)}, \texttt{<critical or not>}\}. You do not have to output all the critical states, but you must give the detailed analysis for the following requirements. Then, you must give the analysis of why our agent lost this episode. Then you must give an analysis of when our racket took great action at those critical states, and how those actions should be rewarded. You have to give a detailed analysis.

\end{tcolorbox}

\par \textbf{(iii) Reward Generation in Pong}

\begin{tcolorbox}[mypromptbox, title=Reward Generation in Pong]
\textbf{Background}\par
It is an $80$×$80$ grid simulation game that mimics table tennis (ping-pong): control the racket to hit the ball sent by the opponent. Whoever fails to catch it loses. The winner’s reward is $1$ point, and the loser’s reward $-1$ point. The game ends when the first person reaches $21$ points.

\textbf{Components of this game}\par
Our racket, the opponent's racket, and the ball move in this $80$×$80$ grid simulation game. Both our racket and the opponent's racket are a $2$×$8$ vector, and ball is a $1$×$2$ vector. Both our racket and the opponent's racket can only move horizontally, which means that their y coordinates will not change. Therefore, we use the tuple \{minimal x, maximal x, minimal y, maximal y\} to denote the positions of rackets. We use the tuple \{x1, x2, y1, y2\} to denote the position of ball. We use the above tuple to denote the movement range of the opponent racket: \{$0$, 7, 8, 9\} $\sim$ \{73, 80, 8, 9\}; We use the above tuple to denote the movement range of our racket: \{0, 7, 70, 71\} $\sim$ \{73, 80, 70, 71\}.

\textbf{Agent objective}\par
We assign our racket with an agent to learn to control how to move in order to hit the ball sent by the opponent and defeat the opponent (first to reach 21 points). We model the game as an MDP: the agent observes its state and takes an action based on its policy and updates its policy based on the reward from the environment. The process repeats until the game ends.

\textbf{How to catch the ball}\par
Only if the ball is close to the 8 coordinate points on the upper surface of the racket. Here is an example: if the ball location is \{52, 69, 53, 69\}, only if the location of our racket is from \{52, 59, 70, 71\} to \{46, 53, 70, 71\}, our racket can catch the ball.

\textbf{Generation instructions}\par
(i) You must assign a numerical reward to guide agent's policy learning based on the agent's location and action and ball location at the timestep " + \texttt{str(timestep)} +". The scale of the reward is from -1 to 1. Your reward assignment must consider the impact degree of the agent's action at this time step: if the agent's action can help it catch the ball, its reward should be positive.

(ii) If agent's action has a bad influence on its future ball catches, its reward should be negative. When you evaluate agent's action, you must infer the ball movement based on ball locations in the continuous time steps to helps you analyze the impact of agent's action to future ball catches.

\textbf{Case Analysis}\par
Here are some analyses of failure reasons in the historical episodes. You have to combine these analyses when you evaluate the agent's action at a certain state to provide a reasonable reward to the agent: 

(i) Several critical states where the agent moved inaccurately, including following the ball too slowly or misjudging position changes...

(iv) The sequence of moves from Timeslot 60 (moving left when it could have been still or moving right) started a pattern of sub-optimal placement of the racket. Timeslot 61/62: The ball, at this heightened y position, needed steady alignment for the racket. The decision at Timeslot 61 to move right further misaligned the racket...

\textbf{Output format}\par
The format of your generated results must be: \{\texttt{reward = <num>, analysis: <your analysis>}\}.

\end{tcolorbox}

\par \textbf{(iv) Critical State Identification and Action Suggestion in Hopper}

\begin{tcolorbox}[mypromptbox, title=Critical State Identification and Action Suggestion in Hopper]
\textbf{Background}\par
The environment is \texttt{Hopper-v4}. It is a continuous control environment based on the MuJoCo physics engine, used for benchmarking reinforcement learning algorithms. It simulates a planar robot (essentially a one-legged creature) trying to hop forward without falling. We make the Hopper hop forward as far as possible while maintaining balance. We model the game as an MDP: the agent observes its state and takes an action based on its policy and updates its policy based on the reward from the environment. The process repeats until the game ends.

\textbf{State space}\par
The state is a 11-dimensional continuous vector capturing the Hopper’s physical condition: (1) torso horizontal position (meters), (2) torso vertical position (used to detect falling) (meters), (3) torso pitch (radians), (4) torso horizontal velocity (m/s), (5) torso vertical velocity (m/s), (6) torso pitch velocity (rad/s), (7) hip joint angle (radians), (8) hip joint angular velocity (rad/s), (9) knee joint angle (radians), (10) knee joint angular velocity (rad/s), (11) ankle joint angle (radians).

\textbf{Action space}\par
The action is a 3-dimensional continuous vector, representing the torque applied to each joint: (1) Torque applied at the hip joint, ranging from -1 to 1, (2) torque applied at the knee joint, ranging from -1 to 1, (3) torque at the ankle joint, ranging from -1 to 1.

\textbf{Reward Function}\par
The reward at each timestep is composed of three parts: (1) \texttt{Forward reward} (primary objective): Reward proportional to the hopper’s horizontal velocity; (2) \texttt{Alive bonus} (survival incentive): A small positive constant reward every timestep the Hopper remains standing (e.g., +1.0), (3) \texttt{Control cost} (penalization): A penalty proportional to the magnitude of the actions, encouraging smoother control. The \texttt{environment reward} = \texttt{forward velocity} + \texttt{alive bonus} - 0.001 * \texttt{control cost}.

\textbf{Termination conditions}\par
The episode ends if any of the following occur: (1) Falling: torso vertical position falls below 0.7 meters, (2) Large Pitch Angle: Absolute torso pitch angle exceeds ~36 degrees (abs(pitch) > 0.628 radians), (3) Numerical Instability: NaN or Inf values appear in the state.

\textbf{Identification instructions}\par
We give you a piece of the agent's trajectory from timeslot " + \texttt{str(begin)} + " to timeslot " + \texttt{str(end-1)} + " in the episode " + \texttt{str(e)} + ". 

(i) Your goal is to identify all the critical states in this time interval. How to identify a state is critical or not is to use reasoning to determine whether the action taken by the agent in each state has an impact on future rewards, that is, whether it affects the agent's ability to maximize the future environment reward. Here is an example: if the robot agent's action in this state will lead to termination in the future, the state is critical. If a robot agent's action in this state greatly impacts its future reward in the further states, it is critical. You must use the robot agent's movement trajectories we provide to analyze how the robot agent's actions in states impact its movement to indirectly impact its reward, so that you can identify all the critical states in the agent's historical trajectories. 

(ii) Your second goal is that after you identify if states are critical or not in this time interval, you must infer if the robot agent's actions are correct at those critical states and give corrected actions if needed. For example, if a robot agent's action in this state will lead to termination in the future, its action is incorrect and needs to be corrected. If a robot agent's action in this state greatly impacts its future reward in future states, its action is incorrect and needs to be corrected. To correct the robot agent's action at each critical state, you must output the corrected actions following the action space in this environment: (Torque applied at the hip joint, torque applied at the knee joint, torque at the ankle joint), ranging from -1 to 1.

\textbf{Output format}\par
The format of your identified results must be: \{\texttt{timeslot str(begin)}, \texttt{<critical or not>}, \texttt{<corrected actions if agents' actions are unproper>}, \texttt{<explanation>}\}, \{\texttt{timeslot str(begin+1)}, \texttt{<critical or not>}, \texttt{<corrected actions if agents' actions are unproper>}, \texttt{<explanation>}\}, ..., \{\texttt{timeslot str(end-1)}, \texttt{<critical or not>}, \texttt{<corrected actions if agents' actions are unproper>}, \texttt{<explanation>}\}.
\end{tcolorbox}

\par \textbf{(v) Reward Generation in Hopper}

\begin{tcolorbox}[mypromptbox, title=Reward Generation in Hopper]
\textbf{Background}\par
The environment is \texttt{Hopper-v4}. It is a continuous control environment based on the MuJoCo physics engine, used for benchmarking reinforcement learning algorithms. It simulates a planar robot (essentially a one-legged creature) trying to hop forward without falling. We make the Hopper hop forward as far as possible while maintaining balance. We model the game as an MDP: the agent observes its state and takes an action based on its policy and updates its policy based on the reward from the environment. The process repeats until the game ends.

\textbf{State space}\par
The state is a 11-dimensional continuous vector capturing the Hopper’s physical condition: (1) torso horizontal position (meters), (2) torso vertical position (used to detect falling) (meters), (3) torso pitch (radians), (4) torso horizontal velocity (m/s), (5) torso vertical velocity (m/s), (6) torso pitch velocity (rad/s), (7) hip joint angle (radians), (8) hip joint angular velocity (rad/s), (9) knee joint angle (radians), (10) knee joint angular velocity (rad/s), (11) ankle joint angle (radians).

\textbf{Action space}\par
The action is a 3-dimensional continuous vector, representing the torque applied to each joint: (1) Torque applied at the hip joint, ranging from -1 to 1, (2) torque applied at the knee joint, ranging from -1 to 1, (3) torque at the ankle joint, ranging from -1 to 1.

\textbf{Reward Function}\par
The reward at each timestep is composed of three parts: (1) \texttt{Forward reward} (primary objective): Reward proportional to the hopper’s horizontal velocity; (2) \texttt{Alive bonus} (survival incentive): A small positive constant reward every timestep the Hopper remains standing (e.g., +1.0), (3) \texttt{Control cost} (penalization): A penalty proportional to the magnitude of the actions, encouraging smoother control. The \texttt{environment reward} = \texttt{forward velocity} + \texttt{alive bonus} - 0.001 * \texttt{control cost}.

\textbf{Termination conditions}\par
The episode ends if any of the following occur: (1) Falling: torso vertical position falls below 0.7 meters, (2) Large Pitch Angle: Absolute torso pitch angle exceeds ~36 degrees (abs(pitch) > 0.628 radians), (3) Numerical Instability: NaN or Inf values appear in the state.

\textbf{Generation instructions}\par
We give you a piece of the agent's trajectory from timeslot " + \texttt{str(begin)} + " to timeslot " + \texttt{str(end-1)} + " in the episode " + \texttt{str(e)} + ". Your goal is to assign a numerical reward to guide the robot agent's policy learning based on its state and action at timeslot " + str(timestep) + ". The scale of the reward is from -1 to 1. Your reward assignment must consider the impact degree of the agent's action at this time step. For example, if the robot agent's action in this state will lead to termination in the future, the reward should be negative. If the robot agent's action in this state greatly reduces its future reward, the reward should be negative; vice versa.

\textbf{Case Analysis}\par
Here are some analyses about effective or sub-optimal actions for the agent. You must combine these analyses when you evaluate the agent's action at timeslot " + \texttt{str(timestep)} + " to provide a reasonable reward to the agent…

\textbf{Output format}\par
The format of your identified results must be: \{\texttt{reward = <num>, analysis: <your analysis>}\}.
\end{tcolorbox}

\section*{C. LLM-generated Explanation}
\label{appendix:C}
\par Here we list examples of LLM-generated explanations in this work.
\par \textbf{(i) An example of reward generation from LLM}

\begin{tcolorbox}[mypromptbox, title=An example of reward generation from LLM]
First, let's clarify the ball movement. From timestep 3 (y=40) to timestep 4 (y=36) to timestep 5 (y=32), the ball's y-coordinate is decreasing. According to your description, this means the ball is moving to the opponent side, not our side. Thus, catching the ball is not an imminent need at this step. However, key parts of your historic failure analysis suggest that suboptimal racket movements—especially unnecessary shifts—can misalign the racket for future returns, notably when the ball will soon move back towards your side.

\textbf{Analysis of Agent’s Action:} \par
- At timestep 4, the agent takes \texttt{move right} from position {12, 19, 70, 71} (goes to {10, 17, 70, 71} in timestep 5 due to the game’s movement rule). 
- The ball, across all timesteps provided, is far from the agent's racket zone (ball x in [74,75], [70,71], [66,67], which is all far right versus agent's x in [12,19], [10,17]). Thus, the agent is not yet in a threatening situation, but its positioning in the horizontal axis becomes relevant once the ball heads back towards you.
- Your failure analysis repeatedly notes that **uncritical, inefficient, or misjudged actions away from likely ball trajectories set up later failures, especially as the ball comes closer.

\textbf{Evaluating the Agent's Action}: \par
- The agent \texttt{moves right} but ends up at {10, 17, 70, 71} (a LEFT shift, based on your movement rule! "\texttt{move right}" subtracts 2 from min/max x). This further moves the racket even further away from the ball (which is at x=66,67 at timestep 5; agent at x=10,17). This is a clear suboptimal move, even though this isn't an imminent catch/touch step.
- This action does not support efficient alignment for the next required catch, and directly reflects the critical failures you described: missed anticipation and preparatory positioning lead to breakdowns later as the ball heads back.

\textbf{Reward Justification:} \par
- Since the action does not immediately result in a failed catch (ball is moving away), but the positioning is actively worsened for future ball returns (and your historic failures penalize just such inefficient transitions), the action’s impact is significantly negative.
- Since no immediate failure happens but the future catch likelihood decreases (and prepping/aligning for returns is essential), this step is proactively penalized, but not with maximal severity.

\textbf{Final decision}: \par 

\texttt{reward} = -0.8  

The agent’s action further misaligns itself despite the ball not moving towards it, clearly reducing future success odds and matching failure patterns in historical analysis.

\end{tcolorbox}

\par \textbf{(ii) Potential improvement strategies based on case analysis from LLM}

\begin{tcolorbox}[mypromptbox, title=Potential improvement strategies based on case analysis from LLM]

\textbf{1. State Space Exploration:} \par
The agent needs to better learn from critical moments where its position affects its ability to intercept the ball. We should introduce a mechanism to reinforce learning during critical times while undermining actions leading to unfavorable outcomes.

\textbf{2. Action and Reward Policy: } \par
The actions that lead to the ball being missed should carry negative rewards or penalties proportionate to the failure severity. The policy should focus on maintaining positional advantage to intercept the ball efficiently.

\textbf{3. Introduction of Prioritized Experience Replay:} \par
Maintain a buffer of experiences and prioritize replaying experiences where the agent failed to intercept since these are more informative toward improving policies and actions in critical states.

\textbf{4. Adaptive Step Sizes: } \par 
Offer dynamic action adjustments, especially in moments approaching ball interception, to ensure more responsive movement based on the ball's trajectory.

\textbf{5. Improve State Evaluation: } \par
The agent should better differentiate between critical and uncritical states by enhancing its state evaluation function. Implementing a mechanism that prioritizes actions based on proximity to the ball in the x-coordinates can significantly improve action selection during critical moments.

 \textbf{6. Action Quality Updates: } \par
 Update the action-value estimates (Q-values) more aggressively when the agent takes actions in critical states. This can be implemented by increasing the learning rate when the ball is close by, ensuring actions leading to successful interceptions are reinforced, and those leading to misses are penalized.
 
\textbf{7. Reward Shaping: } \par 
Increase reward granularity around the critical states. If the agent successfully aligns with the ball's trajectory, it receives a more structured reward, not just a binary success/failure, to encourage better learning curves and intermediary success tracking.

\end{tcolorbox}

\end{document}